\documentclass[letterpaper, 10 pt, conference]{ieeeconf}  % Comment this line out if you need a4paper

\IEEEoverridecommandlockouts                              % This command is only needed if 
\usepackage{times}
\usepackage{epsfig}
\usepackage{graphicx}
\usepackage{amsmath}
\usepackage{amssymb}
\usepackage{pifont}
\usepackage{multirow}
\usepackage{dsfont}
\usepackage{xspace}
\usepackage{cite}
\usepackage{xcolor}
\usepackage[caption=false]{subfig}
\usepackage{tabulary}
\usepackage[pagebackref=true,breaklinks=true,colorlinks,bookmarks=false]{hyperref}

\makeatletter
\DeclareRobustCommand\onedot{\futurelet\@let@token\@onedot}
\def\@onedot{\ifx\@let@token.\else.\null\fi\xspace}
\def\eg{\emph{e.g}\onedot}

\def\ie{\emph{i.e}\onedot}

\makeatother

\newcommand{\rtwo}[1]{\multirow{2}{*}{#1}}

\newcommand{\cmark}{\ding{51}}
\newcommand{\xmark}{\ding{55}}

\newcolumntype{x}[1]{>{\centering\arraybackslash}p{#1pt}}
\newlength\savewidth
\newcommand{\tablestyle}[2]{\setlength{\tabcolsep}{#1}\renewcommand{\arraystretch}{#2}\centering\footnotesize}

\newcommand{\bin}[1]{{\color{black} #1}}
\newcommand{\mb}[1]{{\color{black} #1}}
\newcommand{\binn}[1]{{\color{black} #1}}

\title{\LARGE \bf
Auto4D: Learning to Label 4D Objects from Sequential Point Clouds
}

\author{Bin Yang$^{1,2}$ \quad Min Bai$^{1,2}$ \quad Ming Liang$^{1}$ \quad Wenyuan Zeng$^{1,2}$ \quad Raquel Urtasun$^{1,2}$ % <-this % stops a space
\thanks{\bin{$^{1}$ Uber Advanced Technologies Group.}}%
\thanks{\bin{$^{2}$ University of Toronto. Correspondence: \texttt{byang@cs.toronto.edu, urtasun@cs.toronto.edu}}}%
}

\begin{document}

\maketitle

%!TEX root = root.tex
\begin{abstract}
In the past few years we have seen great advances in \bin{object perception (particularly in 4D space-time dimensions)} thanks to deep learning methods.
However, they typically rely on large amounts of high-quality labels to achieve good performance, which often require time-consuming and expensive work by human annotators.
To address this we propose an automatic annotation pipeline that generates accurate object trajectories in 3D space (\ie, 4D labels) from LiDAR point clouds.
% Different from previous works that consider single frames at a time, our approach directly operates on sequential point clouds to combine richer object observations.
The key idea is to decompose the 4D object label into two parts: \bin{the object size in 3D that's fixed through time for rigid objects}, and the motion path describing the evolution of the object's pose through time.
\mb{Instead of generating a series of labels in one shot,} \bin{we adopt an iterative refinement process where online generated object detections are tracked through time as the initialization.
Given the cheap but noisy input, our model produces higher quality 4D labels by re-estimating the object size and smoothing the motion path, where the improvement is achieved by exploiting aggregated observations and motion cues over the entire trajectory.}
% More specifically, given a noisy but easy-to-get object track as initialization, our model first estimates the object size from temporally aggregated observations, and then refines its motion path by considering both frame-wise observations as well as temporal motion cues.
We validate the proposed method on a large-scale driving dataset and show \binn{a 25\% reduction of human annotation efforts}. We also showcase the benefits of our approach in the annotator-in-the-loop setting.
\end{abstract}
%!TEX root = root.tex
\section{Introduction}

Self-driving vehicles have the potential to revolutionize transportation. One of the key ingredients of any autonomy system is the ability to perceive the scene and predict how it might unroll in the near future. This is important for the subsequent motion planner to plan a safe and comfortable maneuver towards its goal. 
In recent years, we have seen tremendous advances in 3D perception and motion forecasting  \cite{shi2020pvrcnn,he2020structure,yang2019std,yang20203dssd,waymo2019multinet,casas2020implicit} thanks to the adoption of deep learning.
However, deep learning-based approaches require massive amounts of labeled data. In the context of perception and motion forecasting, annotations are required in the form of accurate 3D bounding boxes over time. 
We refer to these space-time bounding box annotations as 4D labels. 

Unlike annotations in the image domain, creation of 3D bounding box labels using LiDAR points is a complex task due to the sparsity of the observations. The exact extent, size, and pose of an object are often unclear from a single observation. Therefore,  human annotators must ``detect" each object in each frame while considering the additional evidence over the trajectory sequence to help accurately estimate the size and location of each object. This \mb{difficulty} is exacerbated by the ambiguity in visualizing 3D information on a 2D display for the annotator and retrieve 3D input. As such, this process is extremely tedious, time consuming and expensive. 
For example, more than 10 hours are required to accurately annotate a 25 seconds scene at 10Hz. 
There is tremendous value in automating label creation or in creating human-in-the-loop refinement systems that correct automatically created labels.

In recent years, researchers have proposed numerous works for object detection \cite{shi2020pvrcnn,he2020structure,yang2019std,yang20203dssd,hdnet,meyer2019lasernet}. However, most of these works use only data from a single timestep or past measurements from a very short history. A number of works further target the task of associating detections over time as object tracks \cite{liang2020pnpnet,weng2020abmot}. However, these techniques produce bounding boxes that are often not consistent over time in size and often do not construct a smooth, accurate, and realistic trajectory. On the other hand, some existing work have attempted to perform 3D bounding box refinement over time by considering the entire trajectory. For example, \cite{milan2014continuous} pioneered the inclusion of handcrafted 3D motion priors to refine detections. \cite{walsh2020leveraging} attempts to generate automatic 4D annotations by focusing on only static vehicles.
However, because of their limitations in motion modeling capability or scenarios targeted, their application in automatic label generation is limited.

Instead, we propose an auto-labeling model that holistically considers the LiDAR points collected over the entire trajectory of an object \mb{by incorporating reasoning from the annotation process by humans}. In particular, we focus on the 4D annotation of vehicles, as they are often the most numerous traffic participants in a scene \mb{and would result in the greatest savings in cost}. Our technique is inspired by iterative (semi-) automatic label generation and refinement techniques such as \cite{acuna2018polyrnn}. We first leverage a pipeline built on existing object detection and discrete association as a cheap but noisy initialization to isolate the trajectories of individual vehicles; we note that this initialization can also come from coarse and fast human annotation. Then, we apply a two-branch network. The first branch is an object size reasoning model that predicts a single set of bounding box dimensions using the noisy aggregation of LiDAR points over the entire trajectory. The second branch consists of a trajectory refinement model which exploits the size estimate from the first branch and employs a spatial-temporal encoder-decoder to refine the motion path of the vehicle.

\mb{As our goal is to validate the performance of our high precision automatic labeling pipeline, it is critical to use a dataset with highly accurate ground truth annotations.} Existing datasets either have a limited number of object trajectories \cite{kitti} \mb{due to the high cost of annotation}, or \mb{sacrifice} label quality by using sparse annotation and simplistic linear interpolation \cite{nuscenes}. Therefore, we demonstrate the effectiveness of our model on a large-scale driving dataset with annotations of much higher quality than available datasets.
Compared with the noisy initialization, we show that our model can improve the number of precise bounding boxes \bin{($\ge$0.9 IoU)} by 36\%, while the baseline using offline detection and tracking only brings a relative gain of less than 2\%.
In addition, we also demonstrate our model's ability to improve label quality in the annotator-in-the-loop setting.
%!TEX root = root.tex
\section{Related Work}

\paragraph{3D Object Detection}

With the growing interest in autonomous driving, a large number of works have targeted 3D object detection in  recent years. Many techniques operate on LiDAR points as input. Due to the sparsity and unordered nature of the point clouds, some techniques rasterize the points into a perspective view representation \cite{meyer2019lasernet,li2016range3d,chen2017mv3d} and leverage well developed detection schemes for images. To better preserve the 3D distances in the scene, a different line of work discretizes the points into voxels and computes various types of handcrafted \cite{engelcke2017vote3deep,li20173dconv,shi2020pvrcnn} or learned features \cite{zhou2018voxelnet} before applying 3D convolutions. An increasingly popular alternative \cite{faf,yang2018pixor,lang2019pointpillars,hdnet} operates in bird's eye view by encoding the height dimension as a feature for efficient computation. Finally, a number of works propose to directly operate on the point set \cite{yang2019std,qi2018fpointnet,shi2020pointgnn}. However, all these methods exploit only the information captured by the sensors at a single instant, or aggregate information over a very short (0.5s) duration \cite{yang20203dssd,yin2020center}. 
In our setting we target the offline bounding box generation task, where we can leverage temporal information over a much longer period to improve the bounding box quality.

\paragraph{Temporal Object Detection and Tracking}
Given a sequence of observations as input, techniques have been proposed to use information over time to increase accuracy of detections at each instant. 
In the case of images, learned flow is often used to propagate information across frames via warping, upon which the aggregated features are used for more accurate detection \cite{zhu2017flowguided,wang2018motionaware}. Other works propagate information between frames implicitly without computing flow \cite{feichtenhofer2017d2t}. In these works, the typically smooth deformations of the bounding box and its location are implicitly learned and serve to improve the detections. However, due to the perspective transformation in image space,  learning  priors for the deformations is complicated. Other works incorporate priors  in 3D space, often using handcrafted costs as part of an energy minimization framework \cite{milan2014continuous}. A relatively recent technique \cite{strobe} uses implicit memory of the 3D scene to exploit temporal clues while also being efficient. Finally, techniques leveraging deep learning to refine frame-wise object detections using temporal information and 3D motion have been proposed \cite{hu2019jointmonocular,liang2020pnpnet}. In contrast with these techniques, we consider an extended sequence of input data at the same time, while also strictly enforcing a constant size constraint that applies to 3D bounding boxes of non-deformable objects such as vehicles. 

\paragraph{(Semi-) Automatic Labeling}
Given the expense of creating human annotated datasets, a number of works have targeted automatic or user-in-the-loop semi-automatic label generation. In the setting of creating 3D bounding box labels, \cite{meng2020weakly,lee2018leveraging} learn to predict the 3D cuboid from the nearby points around the center with a fixed neighborhood size (\ie, object size prior). \cite{wang2019latte} reasons about the cuboid during the first frame given one-click annotation, and then applies a Kalman filter based object tracker to generate the labels in the following trajectory. 
In contrast, our refinement technique can be initialized using either human input or imperfect detections from off-the-shelf detectors.
\cite{zakharov2020autolabeling} proposes an interesting technique that leverages differentiable 3D shape decoding and rendering to iteratively fit detailed 3D vehicle shapes to observed image and LiDAR evidence to generate automatic 3D labels, but only makes use of single frames, unlike our model. Finally, \cite{walsh2020leveraging} restricts its domain to only consider detector-generated bounding boxes of static vehicles and generates a final label using the weighted sum of all associated detections. On the other hand, our work is able to seamlessly handle both dynamic and static objects through its flexible motion modeling. 
In a related realm, works such as \cite{castrejon2017polyrnn,acuna2018polyrnn,ling2019curvegcn} which target automatic labels in the image domain demonstrate both the  possibility and value in combining human abilities with deep learning to greatly accelerate label generation. We note that our technique is likewise able to refine coarse initial annotations as well as detections from preprocessing to achieve greater label accuracy. 
%!TEX root = root.tex
\section{Learning to Label 4D Objects}

\begin{figure*}[t]
\begin{center}
   \includegraphics[width=1.0\linewidth]{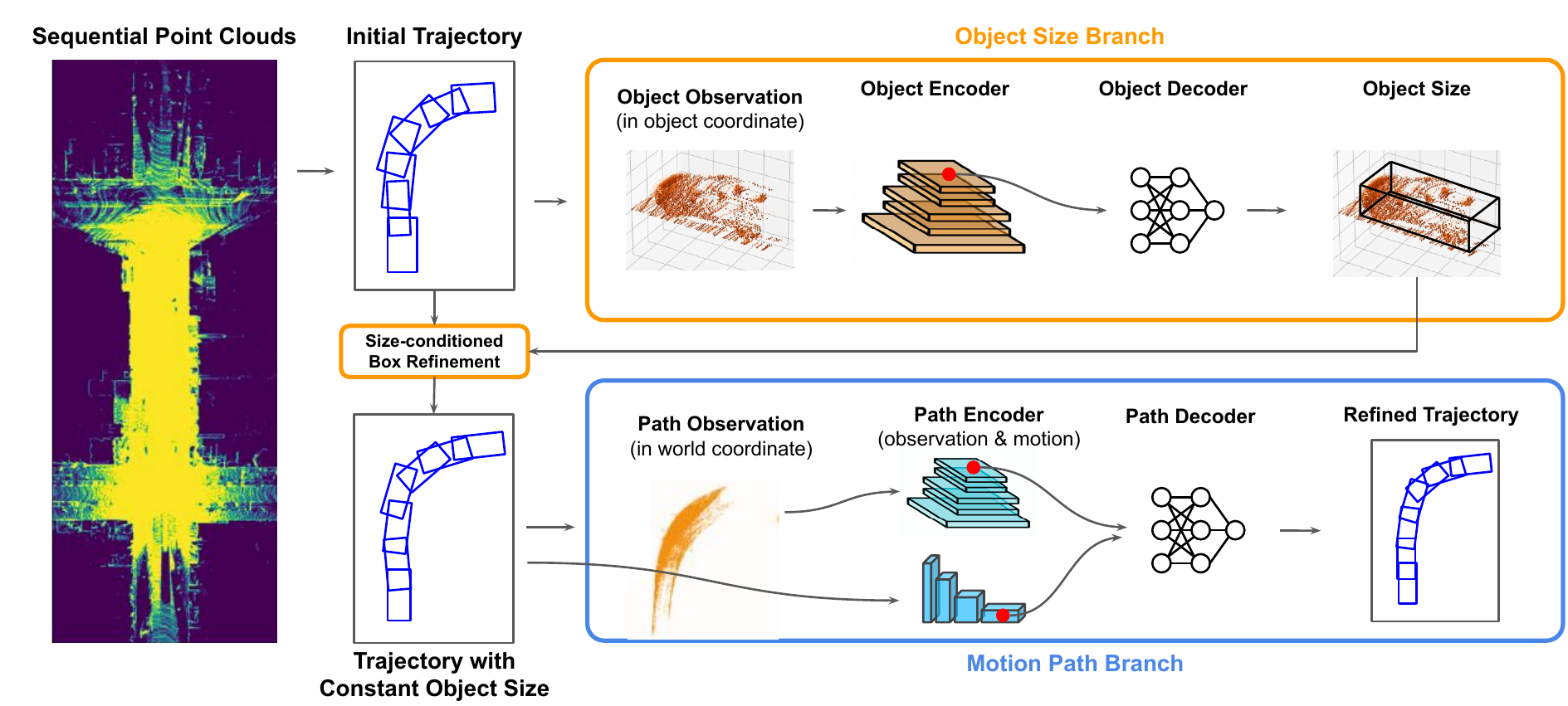}
\end{center}
\vspace{-6mm}
\caption{\textbf{Architecture overview of the proposed model that recovers 4D object labels from sequential point clouds.}}
\vspace{-4mm}
\label{fig:overview}
\end{figure*}

We propose Auto4D, a new automatic labeling approach that recovers  3D object trajectories from sequential point clouds in the context of self-driving.
While previous methods \cite{zakharov2020autolabeling,castrejon2017polyrnn} mostly focus on single-frame object labels either in 2D or 3D, in this paper we aim to generate \textit{temporally consistent} object labels in 3D space (dubbed 4D labels). As vehicles comprise a large portion of dynamic traffic participants in a scene, we focus on the automatic annotation of this class in our work. Note that we can easily extend this work to cover other classes such as pedestrians and bicycles. 
An overview of the proposed method is shown in Fig. \ref{fig:overview}.
% For detailed network architectures and implementation details we refer the readers to the supplementary material.
In this section, we first define the problem we are solving by introducing the input and output representations.
We then explain in turn how we reason about the permanent attribute -- size, and the transient attribute -- pose (or motion path).
Lastly we provide details of the model's training and inference.

\subsection{Input and Output Representations}

\binn{While confident estimates from off-the-shelf models have proven to be useful as pseudo labels \cite{lee2013pseudo}, we observe that there's still a huge gap between these pseudo labels and human annotated labels, especially in terms of localization precision.
To close such gap, our model Auto4D is designed to refine these noisy estimates at trajectory level by reasoning from raw sensor data.}

Auto4D takes as input 3D measurements across time of a scene captured by a LiDAR sensor on top of a self-driving vehicle. Additionally, we use a pre-trained off-the-shelf 3D object detector and tracker to generate the initial object trajectories. 
Auto4D then processes each object trajectory individually, and produces a fixed object size prediction along with a refined motion path. 

In particular, we take the point clouds captured over the whole log sequence to form the observation $\mathbf{x}\in\mathcal{X}$, where $\mathbf{x}$ denotes a set of 4D points (3D position + time). 
We represent these points in the world coordinate system instead of the sensor coordinate system so that we can reason about the object motion independent of the movement of the ego car.

Auto4D extracts object trajectories from the observations in the same world coordinate system. We represent the objects to be labeled with a 2D bounding box in the bird's eye view (BEV) space.
This representation is selected as the typical downstream modules in self-driving (such as the motion planner) reasons in the same BEV space.
\bin{In addition, the position along the height dimension can be determined with heuristics (\eg, ground information from HD maps) in practice.}
We first obtain initial object trajectories by applying a pre-trained off-the-shelf 3D object detector followed with a discrete tracker. 
In particular, we use a voxel based object detector following KITTI state-of-the-art \cite{hdnet}, and a multi-object 3D tracker following \cite{weng20203d}.
As a result, each object trajectory is composed of a set of detections $\mathbf{O}=\{\mathbf{D}_i\}$ where $i$ indexes the frame in which the object is observed. 

Each detection $\mathbf{D_i}=(\mathbf{p},\mathbf{s},t)$ consists of three parts: the object pose $\mathbf{p}=(x,y,\theta)$ indicating the center position and orientation at timestamp $t$, the object size $\mathbf{s}=(w,l)$, and the detection timestamp $t$.
Note that the 3D detector and tracker we used do not exploit the assumption that the object size should be constant. Therefore, the initial estimations for the object's size may vary over the trajectory.

While the initial object trajectory $\mathbf{O}$ is produced in the online setting, our method refines it in the offline setting given the full observation $\mathbf{x}$. 
Our model consists of an object size branch which reasons about a constant and complete object size from $\mathbf{x}$, and a motion path branch which refines the object pose (center position and orientation) along the trajectory in the world coordinate system. Below we explain these two branches in turn.

\subsection{Object Size Branch}
The object size branch aggregates partial observations of the object at different timestamps in the object-relative coordinate system using the bounding boxes provided by the initialization, and produces a single bounding box size for the entire trajectory. 
Over the course of the trajectory, the relative motion between the ego-vehicle and the object allows for observations of the same object from different viewpoints, thus creating a denser and more complete point cloud. 
This richer information allows our model to produce object size estimates with higher accuracy than online detectors that use single (or a small number of) sweeps. 

\paragraph{Object Observation} 
Given the sequential point cloud data $\mathbf{x}$ and initial object trajectory $\mathbf{O}$, we construct a dense point cloud by aggregating the points belonging to the object. 
Specifically, for each detection $\mathbf{D}=(\mathbf{p},\mathbf{s},t)$, we extract points inside the bounding box defined by $\mathbf{D}$ from the corresponding LiDAR sweep. To \mb{be robust to} detection noise, we scale the bounding box size $\mathbf{s}=(w,l)$ by $1.1\times$ when extracting interior points. We then transform these detection-specific points from world coordinates to object coordinates and aggregate these partial observations \mb{over the sequence} \bin{(by aligning the center positions and orientations of detections across all timestamps)}. The resulting object observatio%n $\mathbf{x}_\text{box}^\mathbf{O}$ can be considered as a \textit{noisy} shape reconstruction of the object, where the noise comes from the imperfect detections in the trajectory initialization $\mathbf{O}$.

\paragraph{Object Encoder} 
The object encoder is designed to extract high-resolution spatial features in BEV, which will later be used to reason about the object size. Towards this goal, we project the points in $\mathbf{x}_\text{box}^\mathbf{O}$ to the BEV plane and generate the corresponding pseudo-image in BEV. 
Then, we apply a 2D convolutional neural network (CNN) to effectively enlarge the receptive field and handle variance in object scale, \bin{whose architecture follows the backbone in \cite{liang2020pnpnet} except for the input dimension.}
\begin{equation}
\mathcal{F}_\text{object} = \text{CNN}_\text{box}(\mathbf{x}_\text{box}^\mathbf{O})
\end{equation}
To preserve fine-grained details while being efficient in computation, we generate the BEV pseudo-image by voxelizing the point clouds with a voxel size of $5 \times 5$ cm$^2$. 
Note that the time information in $\mathbf{x}_\text{box}^\mathbf{O}$ is ignored for the purpose of object size estimation. The output of the object encoder is a 3D feature map $\mathcal{F}_\text{object}$ of size $C \times H \times W$, where $H \times W$ are the BEV dimensions.

\paragraph{Object Decoder}
Since the receptive field in $\mathcal{F}_\text{object}$ is large enough to cover the whole object, we simply query the feature at object center from $\mathcal{F}_\text{object}$, and apply a multi-layer perceptron (MLP) to predict the object size.
\begin{equation}
\mathbf{s}' = \text{MLP}_\text{size}(\mathcal{F}_\text{object}(\mathbf{c}))
\end{equation}
The feature query operator is implemented via bilinear interpolation at object center $\mathbf{c}$, \ie, BEV coordinate $(0,0)$.

\begin{figure}[t]
\begin{center}
   \includegraphics[width=.7\linewidth]{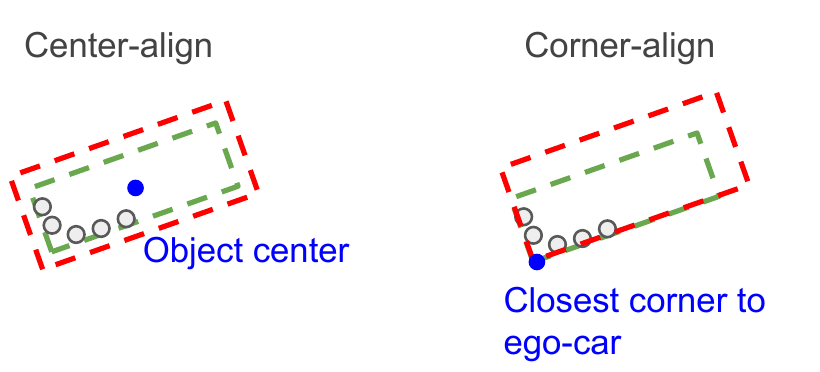}
\end{center}
\vspace{-6mm}
\caption{\textbf{Two different strategies to refine the detection box given the new box size.} The green box is the original box, while the red box is the one with new object size. \bin{We find that aligning the closest corner produces significantly better results as it takes into account different localization uncertainties of visible and occluded corners.}}
\vspace{-4mm}
\label{fig:size_adjust}
\end{figure}

\paragraph{Size-Conditioned Box Refinement}
\bin{By exploiting denser observations over the entire trajectory, the object size branch is able to produce a much more accurate size estimate.
Importantly, this ``noisy shape reconstruction'' works on both moving and static objects regardless of whether or not we have precise localization of the ego-vehicle in the world coordinate, because points are aggregated in the object-relative coordinate.
In the experiments we show that in the case when we have high-precision ego-vehicle localization, we can achieve better size estimation for \textit{static} objects by aggregating points in the \textit{world} coordinate, which bypasses the noise from imperfect detections.}
Given the new size estimate, we update all detections over the trajectory \mb{to leverage the constant size constraint of rigid objects}. One way to achieve this is to update only the width and length of the bounding boxes while retaining the original object center and orientation (``center-aligned''). However, this may not be a good choice in practice, especially on object detections produced from LiDAR data which have biases in their mistakes. This is the case as 3D point observations of vehicles tend to clearly show a subset of the corners and sides of a vehicle, while others are hidden due to occlusion. To address this, we propose to anchor the newly refined box to the corner closest to the ego-vehicle, which is likely the most clearly visible, and adjust the width and length of the bounding box accordingly (``corner-align''). 
We provide a comparison of these two strategies in Fig. \ref{fig:size_adjust}, and find that the ``corner-align'' strategy produces much more accurate trajectories.

\subsection{Motion Path Branch}
\bin{After correcting the object size, we then further smooth the motion path of the object trajectory. 
This is important as the movement of various traffic agents follows certain physical constraints (especially for vehicles in our case), which is under-exploited by online detectors and trackers. 
Some works \cite{kinematic} propose to use a pre-defined kinematic model to capture this, but it doees not generalize well to different vehicles.
In our approach, we adopt a learning based approach that learn from both raw sensor data and sequential object states to predict the refinement to the motion path.}
% The goal of the path branch is to refine the object's motion path conditioned on both the aggregated observations in world coordinates as well as the initial motion path (after size adjustment). 
Relying on the sensor observation helps to better localize the detection at each frame, while exploiting temporal motion information provides additional cues for frames where the object observations are sparse or heavily occluded.

\paragraph{Path Observation}
We generate the aggregated observation by extracting points within the object trajectory $\mathbf{O}=\{\mathbf{D}_i\}$ from the sequential point cloud $\mathbf{x}$ to form $\mathbf{x}_\text{path}^\mathbf{O}$.
The extraction process is largely identical to that used in the object size branch, with the exclusion of the world-to-object coordinate transformation and inclusion of time information in the point clouds. Keeping the 4D points in the world coordinates retains the accurate displacement information for effective motion reasoning.

\paragraph{Path Encoder}
The path encoder is used to extract fine-grained spatial-temporal features from the 4D point clouds $\mathbf{x}_\text{path}^\mathbf{O}$. We observe that vehicles follow largely smooth trajectories in practice, which provides a strong regularization to improve object localization at each time step. As a result, we also explicitly encode the motion feature.
Specifically, we generate a 3D voxel grid by consolidating both the height and time dimension as feature channels \cite{casas2018intentnet}, and apply a 2D CNN to extract multi-scale feature maps \bin{(similar architecture to $\text{CNN}_\text{box}$ with wider channels in the first block to account for additional time input).}
\begin{equation}
\mathcal{F}_\text{path} = \text{CNN}_\text{path}(\mathbf{x}_\text{path}^\mathbf{O})
\end{equation}
The output is a 3D feature map $\mathcal{F}_\text{path}$ with the shape of $C \times H \times W$, where $H \times W$ is the BEV dimension, and $C$ is the channel dimension.
To explicitly model motion, we first extract frame-wise motion features for each detection as the pose displacement from the previous frame.
For the first frame we manually set its motion features to zero.
\begin{equation}
\mathbf{h}_\text{motion}^i = [x^i - x^{i-1}, y^i - y^{i-1}, \theta ^i - \theta ^{i-1}]
\end{equation}
We then concatenate these frame-wise motion features along the time dimension and perform temporal feature extraction with a 1D CNN based UNet \cite{ronneberger2015u}.
\begin{equation}
\mathcal{F}_\text{motion} = \text{UNet}_\text{conv1d}(\mathbf{h}_\text{motion})
\end{equation}

\paragraph{Path Decoder}
We decode the motion path in a convolutional manner. Given the initial detection $\mathcal{D}=(x,y,\theta,w,l,t)$ along the trajectory, we predict its pose refinement $\Delta\mathbf{p}=(\Delta x, \Delta y, \Delta\theta)$ taking into account both the LiDAR voxel feature as well as the motion feature.
\begin{equation}
\Delta\mathbf{p} = \text{MLP}_\text{path}([\mathcal{F}_\text{path}(\mathbf{p}), \mathcal{F}_\text{motion}(t)])
\end{equation}
where $\mathbf{p}=(x,y)$ is the detection center, and $t$ is the detection timestamp. The refined pose $\mathbf{p}'=(x', y', \theta')$ is then computed as follows:
\begin{align}
x' &= (1+\Delta x)*x \\
y' &= (1+\Delta y)*y \\
\theta' &= \theta + \Delta\theta
\end{align}
The voxel feature query $\mathcal{F}_\text{path}(\mathbf{p})$ is implemented as a bilinear interpolation, while the motion feature query $\mathcal{F}_\text{motion}(t)$ is simply done by indexing. 

\subsection{Learning and Inference}
Since the motion path branch (path refinement) depends on the output of the object size branch (box refinement), we train the two branches sequentially. The same loss $\mathbf{L}(\mathbf{D}, \mathbf{G})$ is applied to both branches, which is defined as the Intersection-over-Union (IoU) loss between the estimated and ground-truth bounding boxes. Note that when the IoU is zero (\ie no overlap), the loss falls back to the smooth $\ell_1$ loss between the branch output and ground-truth values.
\begin{equation}
\begin{aligned}
\mathbf{L}(\mathbf{D}, \mathbf{G})= 
\begin{cases}
    -\text{ln}(\text{IoU}(\mathbf{D}, \mathbf{G})), & \text{if } \text{IoU}(\mathbf{D}, \mathbf{G}) > 0\\
    \text{smooth}_{\ell_1}(\mathbf{D}, \mathbf{G}), & \text{otherwise}
\end{cases}
\end{aligned}
\end{equation}
During inference, for each detected object trajectory in the log sequence, we first apply the object size branch to update the frame-wise boxes with a constant size. Then, we apply the motion path branch to the entire trajectory in a sliding window fashion. Note that Auto4D is applicable to both static and moving objects. To avoid small motions for static objects, we add a classification output to the pre-trained object detector to determine whether the object is static or not. If it is a static object, the motion path branch is applied only once on the frame with the highest confidence score.
%!TEX root = root.tex
\section{Experiments}

\bin{We validate the proposed model on a real-world driving dataset with high-quality human annotations. } \mb{We evaluate box agreement at a high threshold (0.9 IoU) to reflect the labeling standard in practice, and approximate the time-savings by computing the percentage of automatically generated bounding boxes that meet this threshold and thus do not need further human input.}
%We evaluate box agreement at a high threshold (0.9 IoU) to reflect the labeling standard in practice, therefore the results illustrate how much human efforts can be saved with our model more realistically.
\bin{As well, we showcase an example of how our model can be used in the annotator-in-the-loop setting by simulating the human input.
Lastly, to provide more insights into how the model works, we perform ablation studies and show qualitative results.}

\subsection{Dataset and Evaluation Metrics}
\bin{High quality labels are important to properly evaluate auto-labeling methods. Public datasets \cite{nuscenes,lyft2019} typically label objects in uniformly placed sparse key-frames (\eg, at 2 Hz) and use linear interpolation to generate per-frame labels (typically at 10 Hz). As a result, the interpolated bounding boxes are imprecise especially when the object changes its velocity.
To address this, we collect a large-scale driving dataset (dubbed Car4D) with LiDAR sensor (Velodyne HDL-64E) data and label with well-trained human experts. In particular, we use adaptively placed key-frames (decided by the annotator) and use the kinematic bicycle model for interpolation to produce higher quality labels.}
We trim 5487 scenes that span 25 seconds each, and create a train/val/test split with 4662/336/489 scenes each.

\bin{The quality of 4D labels consists of various aspects, like box precision, trajectory length, mis-classified objects.
In practice, we observe that drawing a precise box in continuous space takes the most time of a human annotator, as it often requires multiple adjustments.
On the contrary, discrete errors such as false positives, identity switches, trajectory fragmentations can be efficiently corrected by human annotators with a single click.}
Therefore, we designed our model to focus on improving box localization precision, \binn{and focus our evaluation metrics on reducing the continuous errors}.
\bin{To evaluate the bounding box precision, we use IoU in BEV space following common practice in object detection. 
However, instead of taking the 70\% IoU threshold, we focus on 90\% threshold, which reflects the labeling quality of human experts.
% \footnote{For a 2m $\times$ 5m box annotation in BEV, 90\% IoU equals to around 7cm error at the box corners.}
According to \cite{kuznetsova2020open}, the human expert agreement for 2D object boxes in images is 0.88. For LiDAR point clouds, considering the data sparsity and the additional box orientation, the agreement for BEV boxes is likely lower.}

%!TEX root = ../root.tex

\setlength{\tabcolsep}{5pt}
\begin{table}[t]
\caption\rightmark\vspace{-4mm}
{\textbf{Evaluation of 4D labels on Car4D test set.} Our model (Auto4D) is initialized with \textit{online detector + discrete tracker}. At 0.9 IoU criterion, Auto4D saves 25\% of human efforts in correcting poorly localized boxes.}
% \vspace{-2mm}
\begin{center}
\begin{tabular}{l|ccccc}
\hline
\multirow{2}{*}{Method} & \multicolumn{5}{c}{\% of boxes with IoU $\ge$} \\
 & 0.5 & 0.6 & 0.7 & 0.8 & 0.9 \\
\hline\hline
Online detector + & \rtwo{98.8\%} & \rtwo{97.5\%} & \rtwo{94.0\%} & \rtwo{82.2\%} & \rtwo{40.6\%} \\
discrete tracker & & & & \\
\hline
Offline detector + & \rtwo{99.0\%} & \rtwo{97.9\%} & \rtwo{94.7\%} & \rtwo{83.3\%} & \rtwo{41.5\%} \\
discrete tracker & & & & \\
\noalign{\smallskip}
Offline detector + & \rtwo{\textbf{99.5\%}} & \rtwo{\textbf{98.4\%}} & \rtwo{95.0\%} & \rtwo{82.9\%} & \rtwo{41.3\%} \\
disc. \& cont. tracker & & & & \\
\hline
Auto4D (size) & 98.9\% & 97.7\% & 94.8\% & 85.4\% & 49.0\% \\
Auto4D (size + path) & {99.0\%} & {98.0\%} & \textbf{95.6\%} & \textbf{87.9\%} & \textbf{55.3\%} \\
\hline
\end{tabular}
\vspace{-2mm}
\label{tab:result}
\end{center}
\end{table}

% % \setlength{\tabcolsep}{6pt}
% \begin{table*}[t]
% \caption\rightmark\vspace{-4mm}
% {\textbf{Evaluation of 4D object refinement on Car4D test set.} Our approaches are initialized with object trajectories from the online detector and discrete tracker.}
% \begin{center}
% \begin{tabular}{l|ccccc|ccc}
% \hline
% \multirow{2}{*}{Method} & \multicolumn{5}{c|}{Box Recall} & \multicolumn{3}{c}{Corner Recall} \\
%  & $\ge$0.5 IoU & $\ge$0.6 IoU & $\ge$0.7 IoU & $\ge$0.8 IoU & $\ge$0.9 IoU & $\le$20 cm & $\le$10 cm & $\le$5 cm \\
% \hline\hline
% Online detector + & \rtwo{98.8\%} & \rtwo{97.5\%} & \rtwo{94.0\%} & \rtwo{82.2\%} & \rtwo{40.6\%} & \rtwo{60.8\%} & \rtwo{31.5\%} & \rtwo{10.9\%} \\
% discrete tracker & & & & & & & \\
% \hline
% \emph{Offline} detector + & \rtwo{99.0\%} & \rtwo{97.9\%} & \rtwo{94.7\%} & \rtwo{83.3\%} & \rtwo{41.5\%} & \rtwo{61.4\%} & \rtwo{32.0\%} & \rtwo{11.2\%} \\
% discrete tracker & & & & & & & \\
% \noalign{\smallskip}
% \emph{Offline} detector + & \rtwo{\textbf{99.5\%}} & \rtwo{\textbf{98.4\%}} & \rtwo{95.0\%} & \rtwo{82.9\%} & \rtwo{41.3\%} & \rtwo{61.5\%} & \rtwo{32.0\%} & \rtwo{11.3\%} \\
% disc. \& \emph{cont.} tracker & & & & & & & \\
% \hline
% Auto4D (local) & 98.9\% & 97.7\% & 94.8\% & 85.4\% & 49.0\% & 66.4\% & 35.6\% & 12.7\% \\
% Auto4D (local + global) & {99.0\%} & {98.0\%} & \textbf{95.6\%} & \textbf{87.9\%} & \textbf{55.3\%} & \textbf{69.7\%} & \textbf{39.6\%} & \textbf{15.0\%} \\
% \hline
% \end{tabular}
% % \vspace{3mm}
% \label{tab:result}
% \end{center}
% \end{table*}
%!TEX root = ../root.tex

\setlength{\tabcolsep}{6pt}
\begin{table}[t]
\begin{center}
\caption\rightmark\vspace{-2mm}
{\textbf{Evaluation of object size estimation on Car4D val set.}\vspace{2mm}}
\begin{tabular}{l|ccccc}
\hline
\multirow{2}{*}{Method} & \multicolumn{5}{c}{\% of boxes with IoU $\ge$} \\
 & 0.5 & 0.6 & 0.7 & 0.8 & 0.9 \\
\hline\hline
random & \textbf{99.9\%} & 98.8\% & 95.7\% & 84.0\% & 43.2\% \\
mean & 99.8\% & \textbf{99.4\%} & 97.4\% & 89.4\% & 53.0\% \\
median & 99.8\% & \textbf{99.4\%} & 97.4\% & 89.3\% & 54.1\% \\
score & 99.8\% & \textbf{99.4\%} & 97.4\% & 89.1\% & 54.6\% \\
\hline
Auto4D (size) & 99.7\% & 99.3\% & \textbf{97.5\%} & \textbf{91.0\%} & \textbf{61.7\%} \\
\hline
\end{tabular}
\vspace{-4mm}
\label{tab:size2}
\end{center}
\end{table}
%!TEX root = ../root.tex

\begin{table*}[t]
\begin{center}
\caption\rightmark\vspace{-2mm}
{\textbf{Ablation studies on Car4D val set.} Only relative improvements are shown.\vspace{0mm}}

% Sub-tab first row
\subfloat[Ablation on two-branch structure.]{
\tablestyle{2pt}{1.05}
\setlength{\tabcolsep}{6pt}
\begin{tabular}{cc|cc}
\hline
Size & Path & 0.8 IoU & 0.9 IoU \\
\hline
\xmark & \cmark & +2.8\% & +3.7\% \\
\hline
center-align & \xmark & +1.5\% & +3.5\% \\
corner-align & \xmark & +3.3\% & +9.5\% \\
\hline
\cmark & \cmark & \textbf{+6.8\%} & \textbf{+16.3\%} \\
\hline
\end{tabular}}\hspace{5mm}
\subfloat[Ablation on static/moving objects.]{
\tablestyle{2pt}{1.05}
\setlength{\tabcolsep}{6pt}
\begin{tabular}{lc|cc}
\hline
Model & Objects & 0.8 IoU & 0.9 IoU \\
\hline\hline
\multirow{2}{*}{Size} & Static & \textbf{+4.8\%} & \textbf{+12.1\%} \\
 & Moving & {+2.4\%} & {+7.7\%} \\
\hline
\multirow{2}{*}{+Path} & Static & {+2.6\%} & {+4.9\%} \\
 & Moving & \textbf{+4.4\%} & \textbf{+7.9\%} \\
\hline
\end{tabular}}\hspace{5mm}
\subfloat[Ablation on ego-vehicle localization prior (static objects only).]{
\tablestyle{2pt}{1.05}
\setlength{\tabcolsep}{6pt}
\begin{tabular}{c|cc}
\hline
Localization & 0.8 IoU & 0.9 IoU \\
\hline
\xmark & +4.8\% & +12.1\% \\
\cmark & \textbf{+4.9\%} & \textbf{+12.9\%} \\
\hline
\end{tabular}}

\vspace{-6mm}
\label{tab:ablation}
\end{center}
\end{table*}

\subsection{Evaluation of 4D Labels}
\bin{\paragraph{Initialization Model}
The object trajectories are initialized from a state-of-the-art model of \textit{online detector + discrete tracker}, which is a well researched option for obtaining 4D object trajectories. In particular, we train a BEV vehicle detector following the detector architecture of \cite{liang2020pnpnet}. The detector takes past 0.5 second LiDAR data and map information as input. 
For tracking we adopt a distance based tracker similar to \cite{yin2020center}, except that we replace the greedy matching with Hungarian matching \cite{kuhn1955hungarian} for global optimality.}

\paragraph{Baselines}
\bin{Two baselines are proposed in comparison with the object size branch and motion path branch respectively. 
Specifically, while our object size branch exploits observations over the entire trajectory, we compare it with the \textit{offline detector} baseline which takes both past and future 0.5 second LiDAR sweeps as input.
On the other hand, we compare our motion path branch with a \textit{discrete and continuous tracker} where Kalman filtering \cite{kalman1960new} is added to smooth the trajectories.}

\paragraph{Evaluation Results}
\bin{We train the initialization model and the proposed Auto4D model on the Car4D training set, and evaluate on the validation and test sets. This setting mimics the use case when we have labeled a curated dataset, but still have more logs to be labeled. 
The evaluation results on Car4D test set are shown in Table \ref{tab:result}, where we compare the percentage of precise boxes at different IoU criteria.
The initialization model itself already produces pretty accurate object trajectories, with 40.6\% boxes having an IoU with ground-truth greater than 90\%. In other words, these 40.6\% boxes can be used directly as labels if we take 0.9 IoU as labeling standard.
The two baselines further improves the box precision but only to a limited extent. 
Specifically, using offline detector produces more precise boxes at all criteria due to richer observation, with 0.9\% increase in 0.9 IoU.
Adding Kalman filtering produces more boxes at 0.5 and 0.6 IoUs by smoothing the trajectory. However, there's no gain at higher standards as this method does not take raw sensor observation into account.
In contrast, our approach achieves a remarkable improvement. 
Specifically, the object size branch and the motion path branch increase the number of boxes at 0.9 IoU by 8.4\% and 6.3\% respectively. In total, this equals to around 25\%
reduction of human efforts in correcting poorly localized boxes ($<$ 0.9 IoU).}

\bin{\subsection{Evaluation of Size Refinement}
\paragraph{Baselines}
Here we evaluate the impact of the object size branch in more detail by comparing with numerous baselines. Given an object trajectory produced by online detector and tracker with non-constant object size, there are many strategies to achieve a better size estimate by exploiting information over the trajectory. The na\"ive baseline is to randomly select the size estimate from one frame, which basically reflects the online detection's performance. A more robust solution is to take the mean or median size over the entire trajectory to reduce the noise due to imperfect detections. Finally, another solution is take detection uncertainty into account, and select the size estimate from the frame where the detector has the highest confidence score.

\paragraph{Evaluation Results}
We compare our object size branch with all 4 baselines and show the evaluation results in Table \ref{tab:size2}. Given the new size estimate we use our proposed size-conditioned box refinement to adjust the bounding box. We show that  simple methods like \textit{median} or \textit{score} can achieve a much better size estimate with 10.9\%/12.4\% increase in the number of very precise boxes ($\ge$ 0.9 IoU) compared to random sampling. On the other hand, our approach almost doubles the gain (18.5\%) by exploiting aggregated raw sensor observations with a learning based model.
}

\subsection{Evaluation with Annotator-in-the-Loop}
\bin{In addition to providing a much better initialization for human annotators, our model can also be used in iterative semi-automatic label generation with annotator in-the-loop providing minimal input. 
\binn{While in practice the annotators may spend most time correcting continuous errors and directly producing more accurate labels, here we showcase an example where the annotators are only allowed to correct one type of discrete errors (trajectory fragmentation). This shows us whether this minimal feedback can indirectly help Auto4D produce more accurate labels.}
% While the output object trajectories of Auto4D may have various discrete errors like false positives, false negatives, trajectory fragmentations, identity switches, we note that these errors are easily fixed by a human reviewer. Here, we showcase an example where trajectory fragmentation errors are fixed by human annotators. 
Specifically, we simulate this human input by linking trajectories together if they belong to the same object instance. This represents minimal manual work, as only two relatively imprecise clicks are required in practice. After fixing the fragmentation errors, we feed the new trajectories to our Auto4D model again for a second iteration of refinement. In this simple example of annotator in-the-loop, we show that after the second iteration we can still achieve 1.0\% increase in the number of very precise boxes ($\ge$ 0.9 IoU). Note that this is achieved without re-training a new Auto4D model for the second iteration. We expect further improvements from our model with more iterations if other types of errors are fixed by human input.}

\begin{figure*}[t]
\begin{center}
   \includegraphics[width=0.19\linewidth]{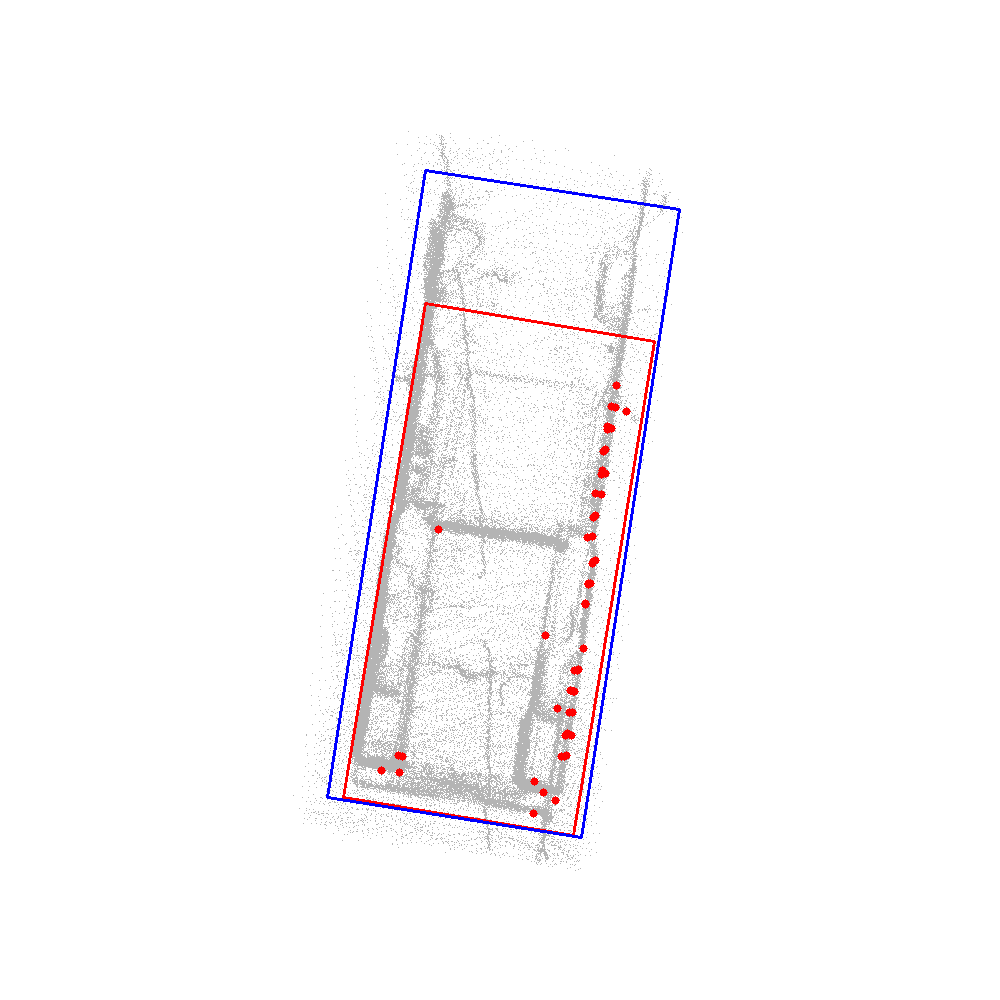}\includegraphics[width=0.19\linewidth]{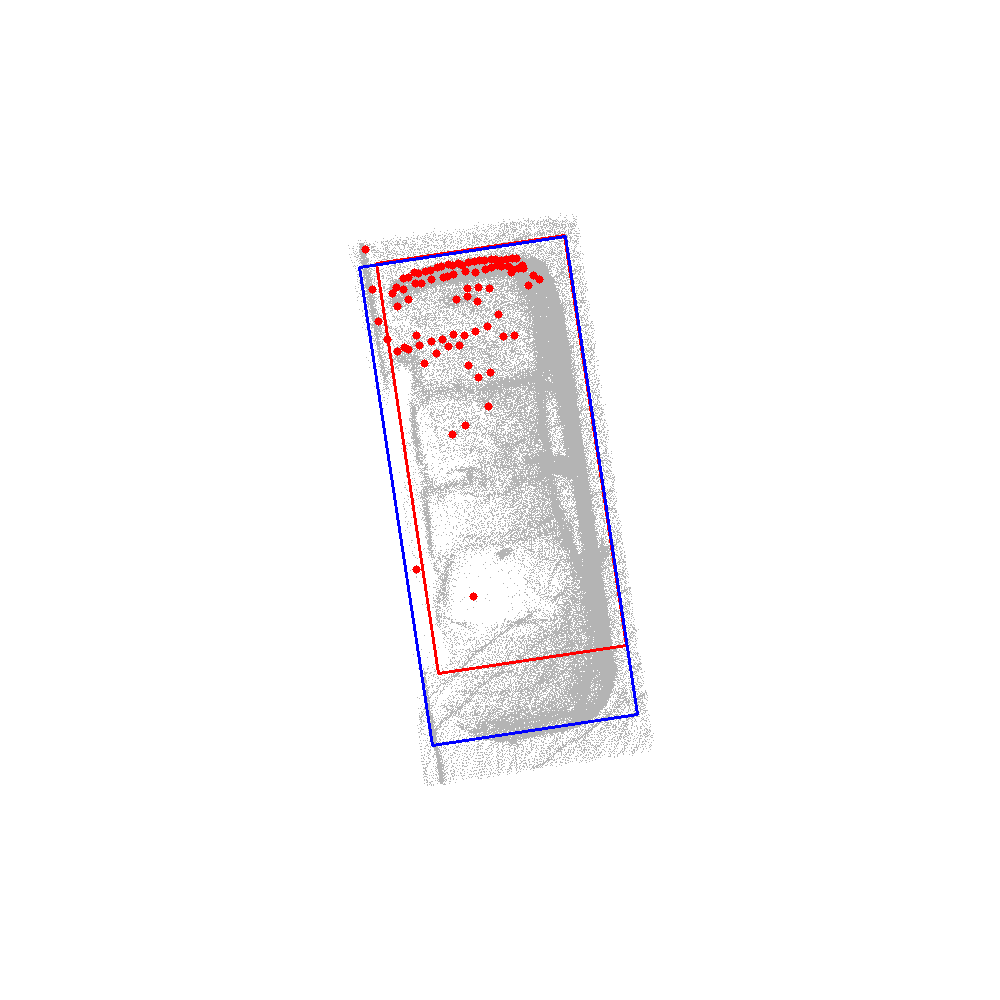}\includegraphics[width=0.19\linewidth]{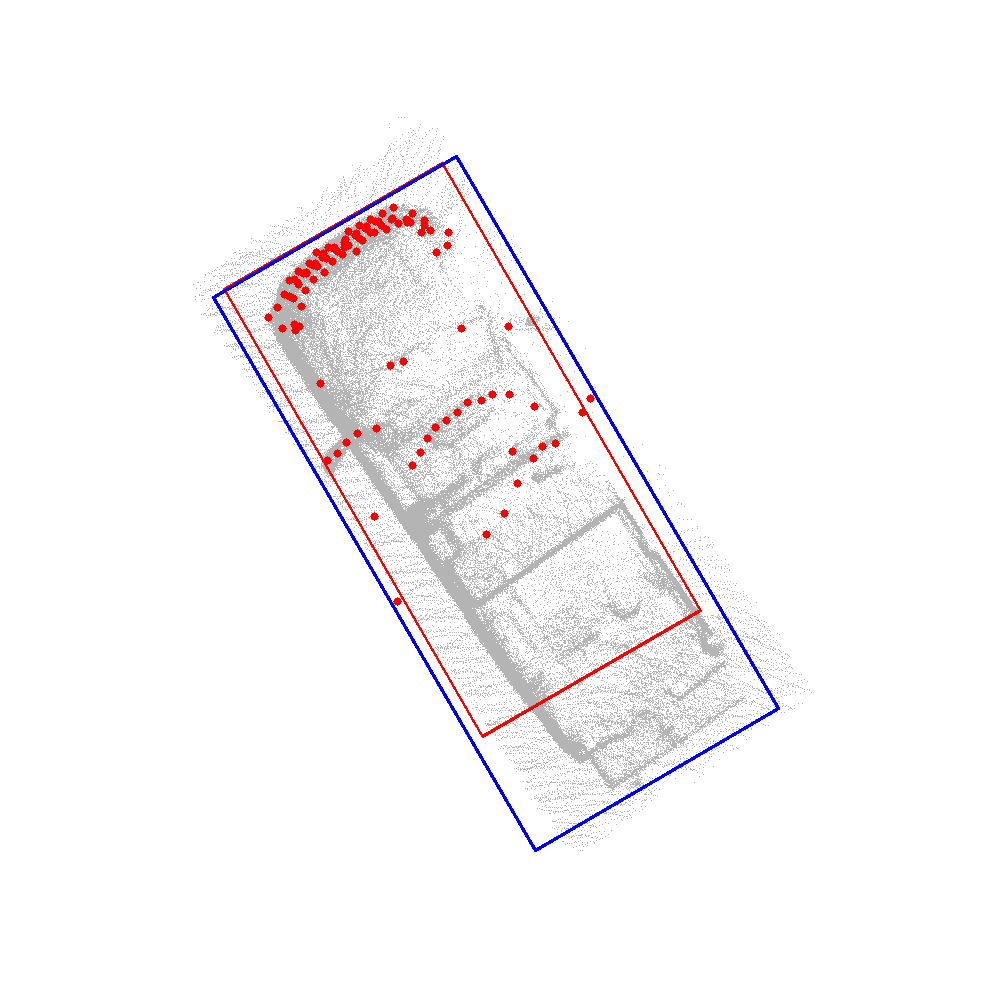}\includegraphics[width=0.19\linewidth]{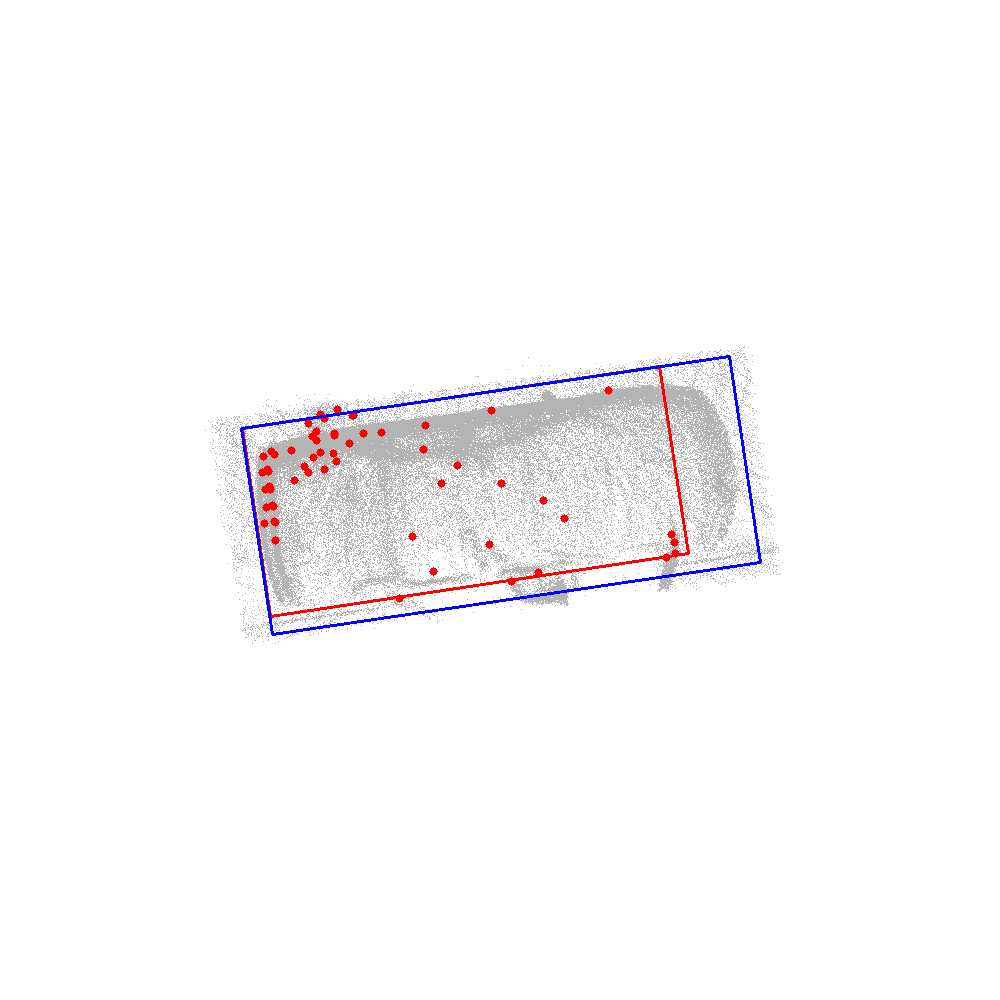}\includegraphics[width=0.19\linewidth]{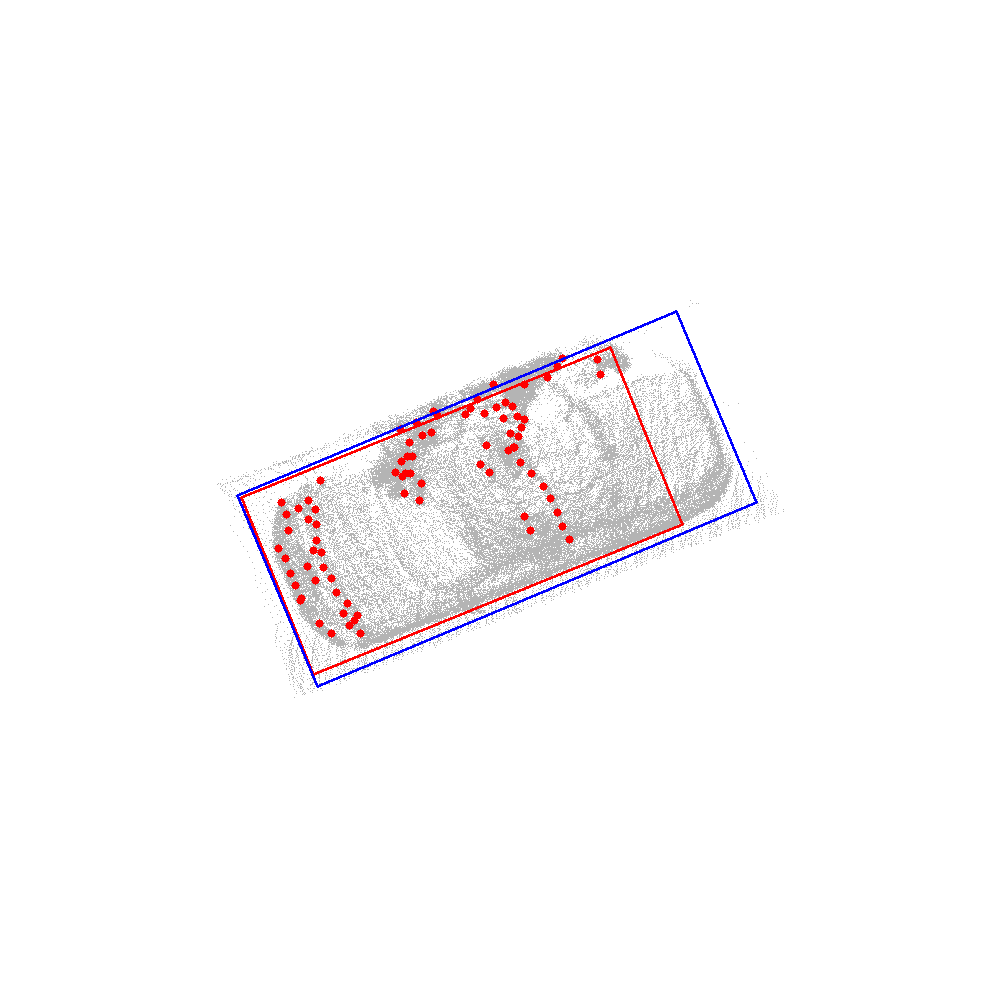}
   \includegraphics[width=0.19\linewidth]{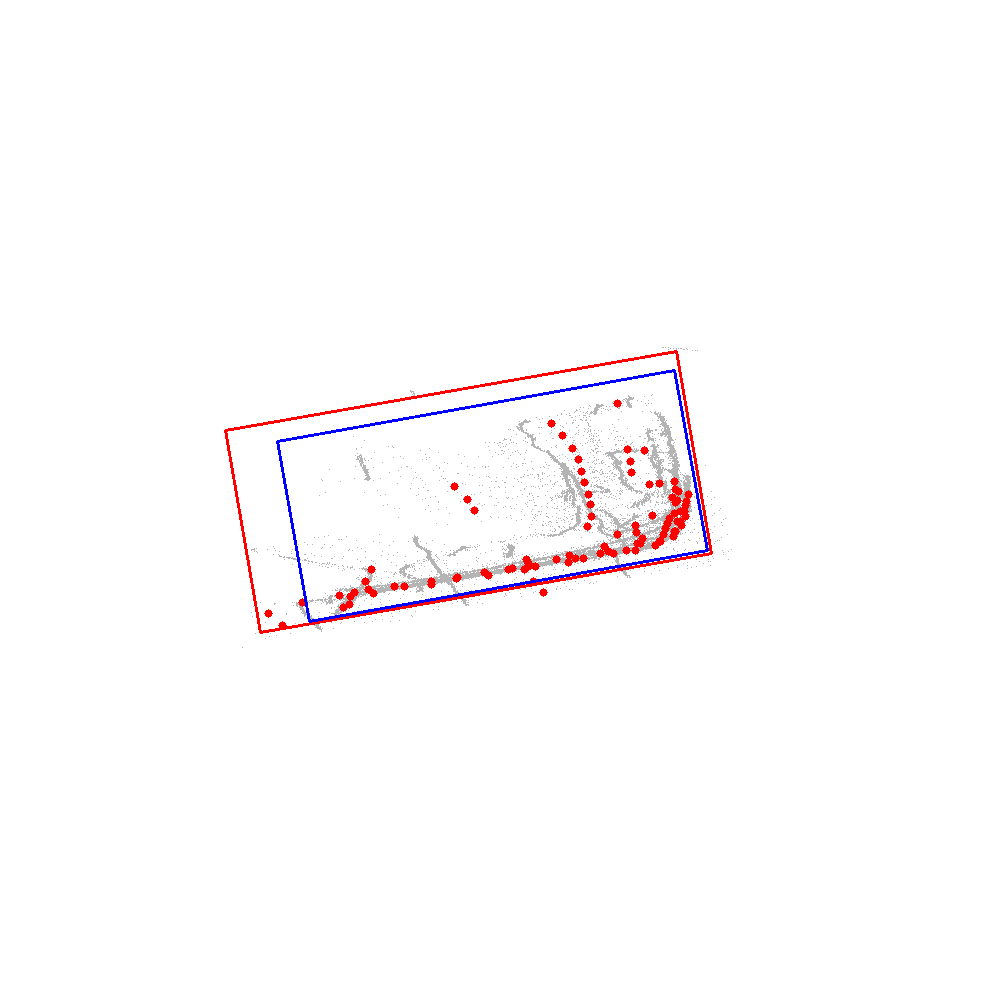}\includegraphics[width=0.19\linewidth]{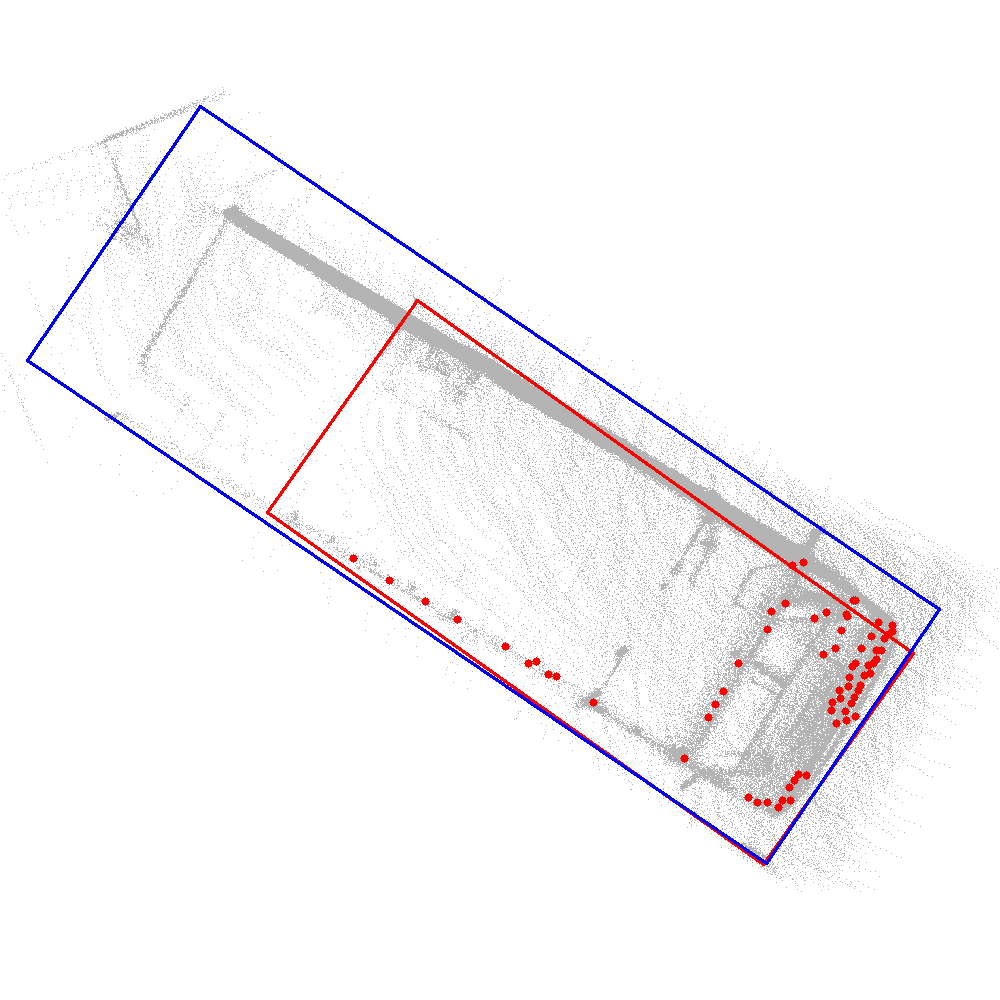}\includegraphics[width=0.19\linewidth]{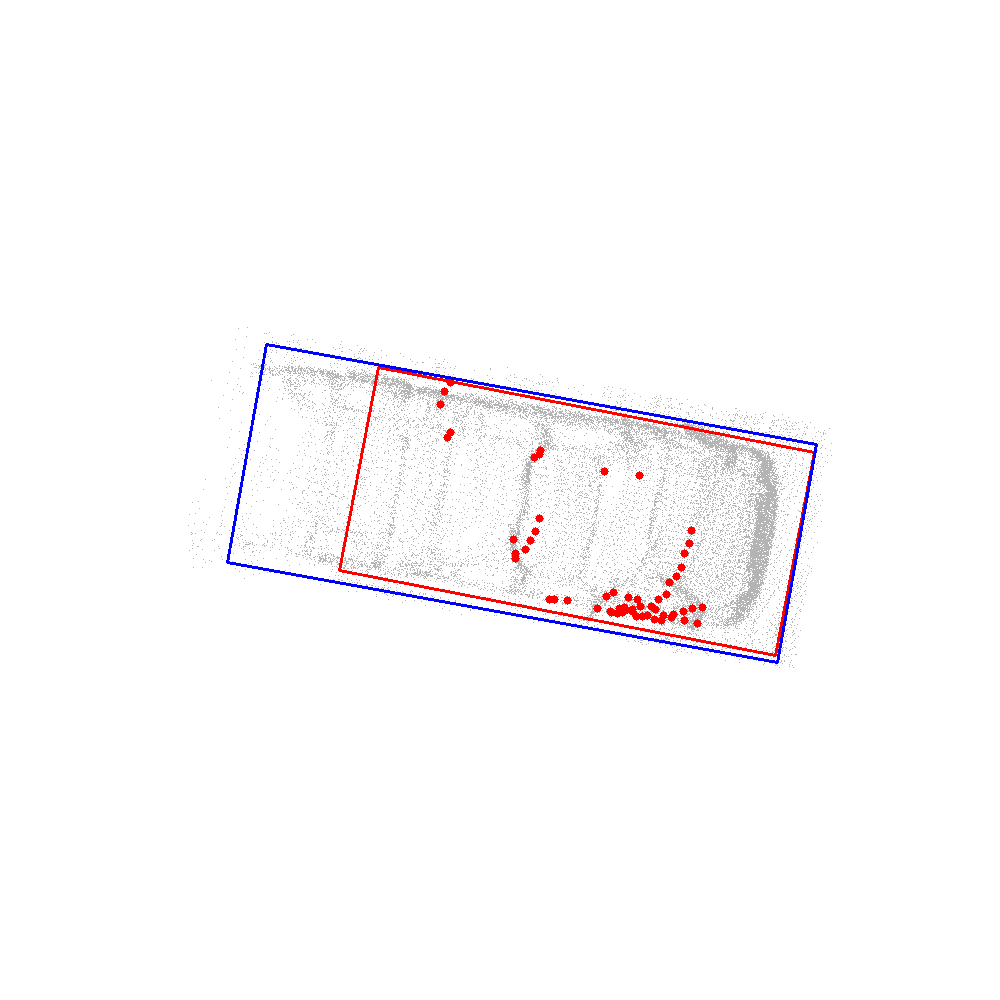}\includegraphics[width=0.19\linewidth]{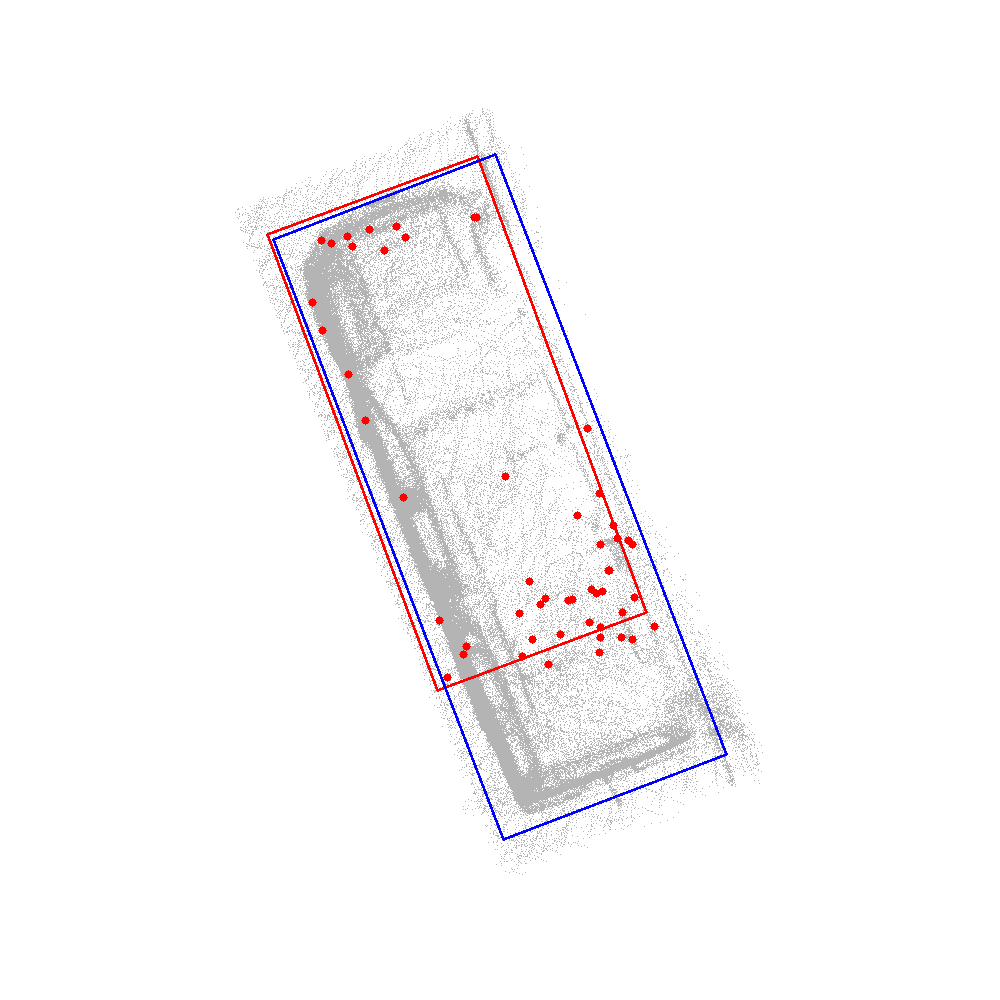}\includegraphics[width=0.19\linewidth]{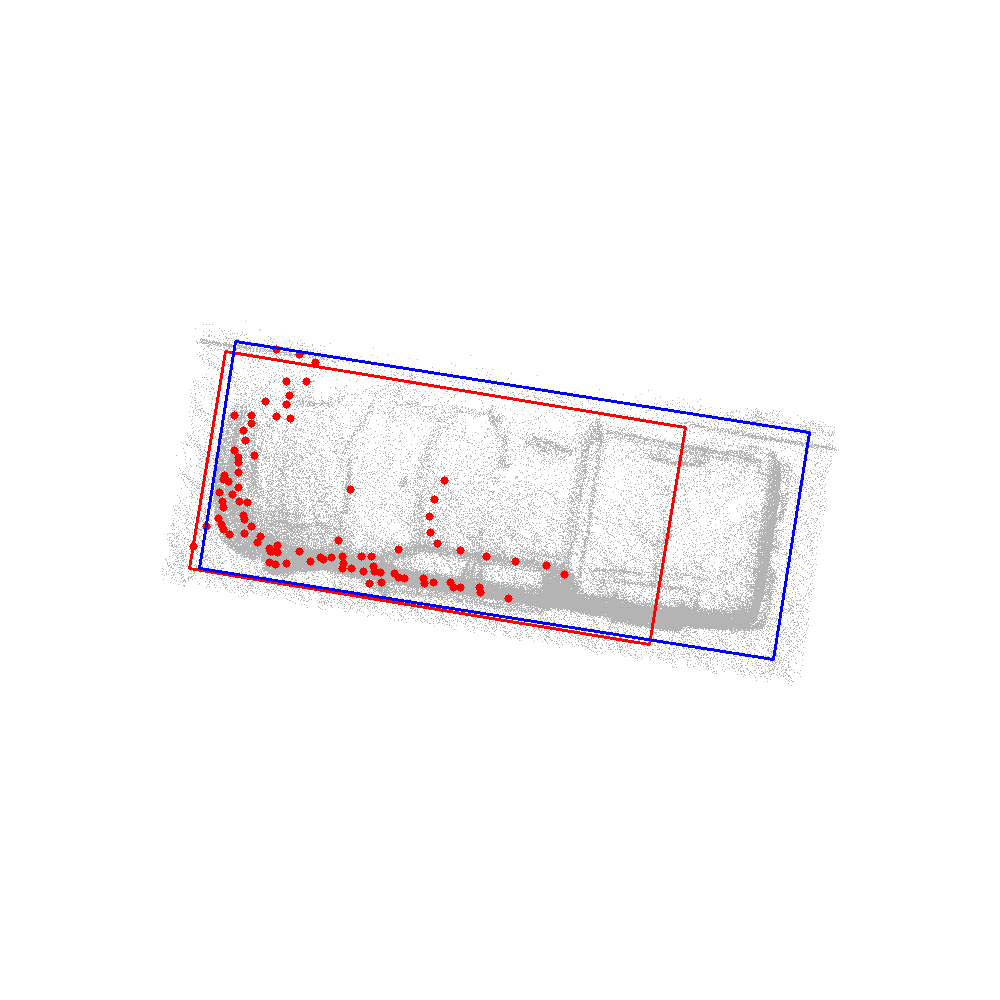}
\end{center}
\vspace{-6mm}
   \caption{\textbf{Qualitative comparison of our refined labels (in {\color{blue}blue}) and the initial detections (in {\text{red}red}).} We show examples of static vehicles only. We draw aggregated LiDAR points in {\color{gray}gray}, and single-sweep LiDAR points corresponding to the detection timestamp in {\color{red}red}.}
\label{fig:demo_size}
% \vspace{-0mm}
\end{figure*}

\begin{figure*}[t]
\begin{center}
   \includegraphics[width=0.9\linewidth]{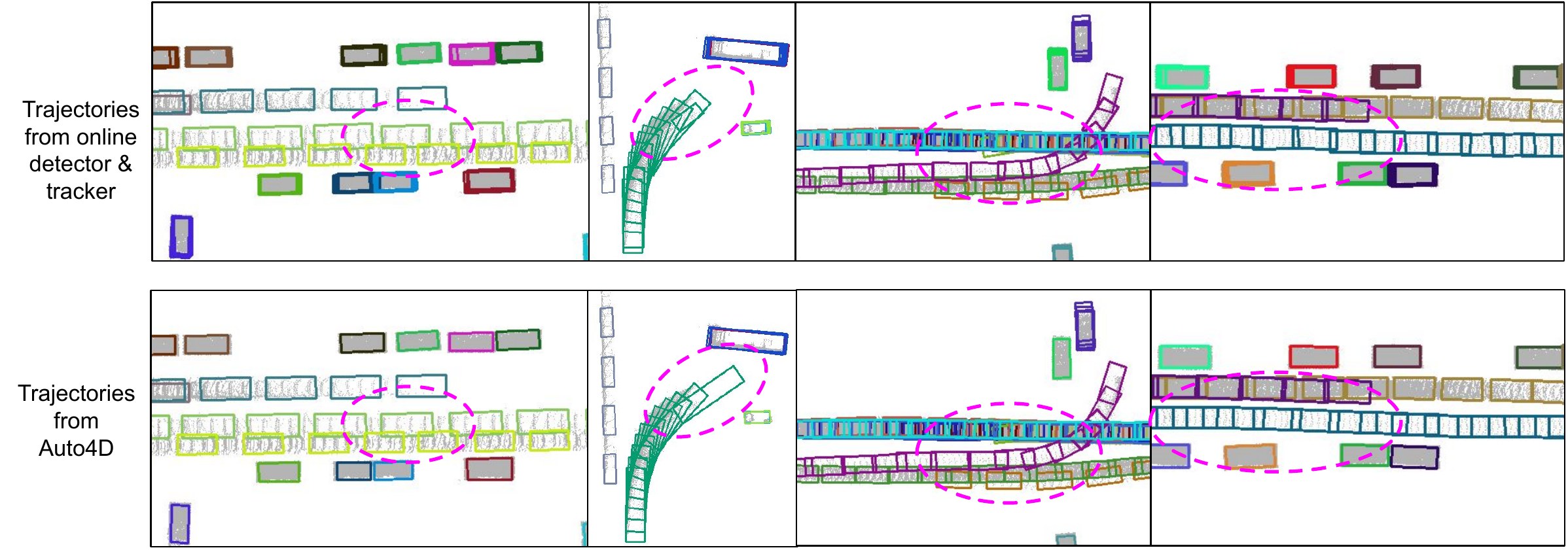}
\end{center}
\vspace{-4mm}
\caption{\textbf{Qualitative comparison of the online detection \& tracking results and Auto4D results.} We draw trajectories (color) together with corresponding LiDAR points ({\color{gray}gray}) in BEV. We draw bounding boxes at 2 Hz to avoid clutter. Significant improvements are highlighted with {\color{magenta}magenta} circles.}
\vspace{-4mm}
\label{fig:demo}
\end{figure*}

\subsection{Ablation Studies}

\bin{We perform several ablation studies to analyze and validate our design decisions, and show the evaluation results on Car4D validation set in Table \ref{tab:ablation}.

\paragraph{Two-Branch Structure}
Auto4D decomposes the object size from the motion path and solves the two components separately. 
An alternative is to optimize both together. To implement this, we remove the object size branch, and add a size output to the motion path branch (therefore no assumption of constant object size over the trajectory). 
As shown in Table \ref{tab:ablation} (a), compared with the two-branch structure (16.3\% gain), this baseline only achieves 3.7\% gain.
We also verify the advantage of the proposed size-conditioned box refinement by comparing the center-align strategy with our corner-align strategy, and show that corner-align strategy works much better (3.5\% versus 9.5\%).

\paragraph{Static vs. Moving Objects}
We show fine-grained evaluation results on static and moving objects in Table \ref{tab:ablation} (b).
With the object size branch alone, we see that more improvements are achieved on static objects. This is because for static objects we are more likely to get observations from multiple views than dynamic objects, and therefore the size estimation from temporally aggregated observation will be more accurate.
When the motion path branch is applied as well, we see more gains coming from moving objects. This is expected, as the motion path branch is designed to exploit both LiDAR observation and temporal motion cues to refine the motion path.

\paragraph{Ego-Vehicle Localization Prior}
One advantage of the proposed object size branch is its ability to handle both static and moving objects, as the multi-frame LiDAR points are aggregated in the object-relative coordinate system. This may not be the best solution for size estimation of static objects, since this introduces additional errors from frame-wise detections. In the case where we have high-precision ego-vehicle localization (Car4D dataset has centimeter-level accuracy in localization), we can aggregate multi-frame points of static objects in the world coordinate and get a cleaner object reconstruction. In Table \ref{tab:ablation} (c) we show that using such prior brings an additional performance gain, but to a limited extent. This proves that our object size branch itself can learn to recover accurate object size even from noisy object reconstructions.}

\subsection{Qualitative Results}
\bin{We show qualitative results on object detections and trajectories in Fig. \ref{fig:demo_size} and Fig. \ref{fig:demo} respectively, which qualitatively shows our model's ability to estimate the full object size and smooth the motion path.
In particular, in Fig. \ref{fig:demo_size} we see that online LiDAR based object detectors often have difficulty estimating the full extent of the object due to partial observation. In contrast, our approach addresses this by exploiting richer observations over the entire trajectory.
In Fig. \ref{fig:demo}, we compare Auto4D's results with the input noisy trajectories. We find that Auto4D produces more accurate bounding boxes for static objects without small jitters. For moving objects, thanks to the fixed size constraint, we are able to correctly estimate the object's location even when it is far away from the ego-vehicle and the observation is very sparse. Overall the generated trajectories tend to be smoother and more natural. 
We refer readers to the supplementary video for more trajectory-level results.}
%!TEX root = root.tex
\section{Conclusion}
While previous automatic labeling methods mainly focus on generating 2D or 3D labels at frame level, in this paper we take one step forward by using extended temporal information to produce smooth and consistent object trajectories over time.
Towards this goal we propose to decompose the 4D label into two components: object size and motion path. The object size is fixed across time, while the motion path describes how the object pose (center position and orientation) changes over time.
Our model works by taking imperfect but cheap object trajectories as input, and learns to refine them (size \& path) by exploiting spatial-temporal information over the full trajectory.
We validate the approach on a large-scale driving dataset with high-quality annotations, and show significant improvement over various baselines.
% Because our model follows the refinement paradigm, it is also suitable for the annotator-in-the-loop framework.
In the future, we plan to extend the approach to incorporate additional information sources such as camera images.

{\small
\bibliographystyle{IEEEtran}
\bibliography{egbib}
}

\end{document}